%% file: main.tex
\documentclass[10pt, conference, letterpaper]{IEEEtran}
\usepackage{amsmath,amsfonts}
\usepackage{algorithm}
\usepackage{algpseudocode}

\usepackage{array}
\usepackage{textcomp}
\usepackage{stfloats}
\usepackage{verbatim}
\usepackage{graphicx}
\usepackage{cite}
\usepackage{multirow}
\usepackage{subcaption} 
\usepackage{caption}    
\usepackage{float}
\usepackage{placeins}
\usepackage{wrapfig}
\usepackage{makecell}
\usepackage{tikz}

\def\BibTeX{{\rm B\kern-.05em{\sc i\kern-.025em b}\kern-.08em
    T\kern-.1667em\lower.7ex\hbox{E}\kern-.125emX}}

\usepackage{titlesec}
\titlespacing*{\section}{0pt}{*0.8}{*0.8}
\titlespacing*{\subsection}{0pt}{*0.6}{*0.6}
\titlespacing*{\subsubsection}{0pt}{*0.5}{*0.5}

\begin{document}
\title{Federated Learning for Efficient Condition Monitoring and Anomaly Detection in Industrial Cyber-Physical Systems}

\author{\IEEEauthorblockN{William Marfo, Deepak K. Tosh, Shirley V. Moore} \\
\IEEEauthorblockA{\textit{Department of Computer Science}, \textit{University of Texas at El Paso}, El Paso, USA \\
\ wmarfo@miners.utep.edu, dktosh@utep.edu, svmoore@utep.edu
}}

\maketitle


\input{01-abstract}
\input{10-introduction}

\input{30-related-work}
\input{40-methods}

\input{50-results}
\input{55-discussion}
\input{60-conclusion}

\small
\bibliographystyle{IEEEtran}
\bibliography{70-bibliography}

\vfill

\end{document}

%% file: 01-abstract.tex
\begin{abstract}
Detecting and localizing anomalies in cyber-physical systems (CPS) has become increasingly challenging as systems grow in complexity, particularly due to varying sensor reliability and node failures in distributed environments. While federated learning (FL) offers a foundation for distributed model training, existing approaches lack mechanisms to handle these CPS-specific challenges. This paper presents an enhanced FL framework that introduces three key innovations: adaptive model aggregation based on sensor reliability, dynamic node selection for resource optimization, and Weibull-based checkpointing for fault tolerance. Our framework enables reliable condition monitoring while addressing the computational and reliability challenges of industrial CPS deployments. Experiments on NASA Bearing and Hydraulic System Datasets demonstrate superior performance over state-of-the-art FL methods, achieving 99.5\% AUC-ROC in anomaly detection and maintaining accuracy under node failures. Statistical validation using Mann-Whitney (\(\mathrm{U}\)) test confirms significant improvements $(p < 0.05)$ in both detection accuracy and computational efficiency across diverse operational scenarios.

\end{abstract}

\footnote{This material is based upon work supported by the United States Department of Energy’s (DOE) Office of Fossil Energy (FE) Award DE-FE0031744.}


%% file: 10-introduction.tex
\section{Introduction} 
\label{sec:Introduction}

The proliferation of Internet of Things (IoT) devices and autonomous systems within cyber-physical systems (CPS) has heightened the importance of anomaly detection and localization in industrial component health monitoring \cite{rumesh2024federated}. Modern CPS, encompassing smart grids and industrial control systems, generate vast amounts of data from numerous sensors and actuators \cite{shrestha2024anomaly}. With millions of machine failures occurring globally each year, the financial losses due to downtime and repairs are substantial. Early detection and precise localization of sensor anomalies are crucial to minimize these losses and prevent cascading failures.

Traditional machine learning (ML) approaches, while extensively applied to detect anomalies in CPS \cite{xu2024end,marfo2022condition,gaba2024innovative}, face significant challenges in these complex environments. They struggle with the computational burden of analyzing large-scale sensor data from geographically dispersed locations due to the increased need for coordination, communication, and synchronization across nodes. Moreover, these approaches often prioritize detection over precise anomaly localization—identifying the specific components or sensors responsible for the anomalies—limiting their effectiveness in identifying specific problem areas \cite{marfo2022condition}.
 Another critical issue is their lack of resilience against system disruptions, such as node failures or faults in learning models, which can compromise overall system reliability. In CPS environments, system disruptions are particularly challenging to manage due to their cascading impact, potentially resulting in prolonged downtime and extensive recovery efforts \cite{benoit2024checkpointing}.

While existing approaches attempt to address these challenges through various distributed learning methods, they often fail to consider the unique characteristics of CPS environments, such as sensor reliability variations, dynamic operational conditions, and the need for continuous monitoring. Additionally, existing methods lack robust mechanisms for handling the inherent uncertainties and failures common in industrial settings. These limitations indicate the need for a comprehensive framework that not only leverages distributed learning but also incorporates CPS-specific optimizations and robust fault tolerance mechanisms.

To address these challenges, we propose an approach leveraging federated learning (FL) for both anomaly detection and localization in CPS. While FL offers a foundation for distributed model training \cite{rumesh2024federated,gaba2024innovative,xu2024end}, basic FL implementations face several limitations in CPS environments, including (1) inability to handle varying sensor reliability and data quality, (2) vulnerability to node failures and subsequent data loss, (3) inefficient resource utilization across heterogeneous nodes, and (4) limited adaptation to dynamic operational conditions.

Our framework enhances traditional FL approaches through specific components, including an adaptive model aggregation strategy that dynamically weights node contributions based on sensor reliability and data quality, a dynamic node selection mechanism optimized for CPS environments that balances computational load and detection accuracy, and an intelligent checkpointing system that predicts and prevents training disruptions while minimizing overhead.

Building upon these FL enhancements, we implement a specialized checkpointing mechanism designed to address the unique challenges of node failures within decentralized CPS. While our enhanced FL approach improves overall system efficiency, the critical nature of industrial monitoring systems requires additional fault tolerance guarantees. Our adaptive checkpointing mechanism ensures continuous operation by predicting potential node failures using operational patterns and historical data, dynamically adjusting checkpointing frequency based on system conditions and criticality, and enabling rapid recovery without compromising model accuracy or training progress.

We aim to answer the following research questions (RQs):

\begin{itemize}
  \item \textbf{RQ1:} How can distributed CPS environments achieve reliable real-time condition monitoring through our enhanced FL framework's novel components?
  \item \textbf{RQ2:} What are the performance advantages of our FL approach in terms of execution time and scalability across different CPS datasets?
  \item \textbf{RQ3:} How does our proposed checkpointing mechanism impact the resilience of the FL system in the presence of node failures?
\end{itemize}

To address these questions, this paper makes the following contributions:

\begin{itemize}
  \item We develop an enhanced FL framework for CPS environments that features: (1) adaptive model aggregation based on sensor reliability, (2) dynamic node selection for resource optimization, and (3) integrated anomaly localization.
  
  \item We validate our approach using NASA Bearing and Hydraulic Systems Datasets, achieving up to 99.5\% AUC-ROC detection accuracy and approximately 2x faster execution compared to FedAvg while maintaining superior performance over ACFL and FedL2P.
  
  \item We design an adaptive checkpointing mechanism using Weibull distribution modeling that ensures fault tolerance with minimal overhead while maintaining model consistency.
\end{itemize}

The remainder of this paper is organized as follows: Section \ref{sec:relatedwork} reviews related work on federated learning for anomaly detection in CPS. Section \ref{sec:framework} presents our proposed framework. Section \ref{sec:Results} provides experimental results. Section \ref{sec:Discussion} discusses the implications of these results and addresses the limitations of our approach. Finally, Section \ref{sec:Conclusion} concludes with key findings and future directions.

%% file: 30-related-work.tex
\section{Related Work} 
\label{sec:relatedwork}

Our research is informed by past work leveraging FL and anomaly detection techniques for enhancing the reliability of CPS. Here, we provide an overview of related work in this area.

Shrestha et al. \cite{shrestha2024anomaly} proposed an anomaly detection framework based on LSTM and autoencoders using FL for smart electric grids. They demonstrated that their approach, which employs Mean Standard Deviation (MSD) and Median Absolute Deviation (MAD) techniques, effectively detects anomalies while preserving data privacy through homomorphic encryption. Their framework achieved a 98\% accuracy with the MSD approach, highlighting the trade-off between privacy, performance, and computation time. Gaba et al. \cite{gaba2024innovative} introduced a vertical federated multi-agent learning framework for CPS. They demonstrated that their approach, which uses synchronous Deep Q-Network (DQN) and Advantage Actor-Critic (A2C) agents, significantly improves cybersecurity by effectively learning optimal policies in both static and dynamic environments. Their findings show substantial improvements over standard methods, with up to 47.26\% higher performance compared to traditional reinforcement learning techniques. Xu et al. \cite{xu2024end} proposed an end-edge collaborative lightweight secure FL (LSFL) architecture for anomaly detection in wireless industrial control systems. They demonstrated that their LSFL approach, which integrates RMS-CNN and adaptive key generation algorithms, achieves over 99\% accuracy while reducing communication costs by up to 89.6\%. This approach effectively balances computation resources, communication costs, and security requirements in FL environments. Rumesh et al. \cite{rumesh2024federated} presented a security architecture for Open Radio Access Networks (O-RAN) utilizing a Network Digital Twin (NDT) framework. They showed that their hierarchical FL-based anomaly detection algorithm accurately identifies anomalous traffic in O-RAN with over 99\% accuracy. This work emphasizes the importance of ML model training in simulated environments before deployment in physical networks, enhancing the security of O-RAN systems. Taheri et al. \cite{taheri2024mitigating} proposed an artificial neural network (ANN)-based adaptation of FL-trust to mitigate cyber anomalies in virtual power plants (VPPs). They demonstrated that their ANN-based approach outperforms traditional FL-trust with a PI controller, particularly in handling non-IID datasets, and effectively mitigates poisoning attacks. Their results underscore the superior accuracy and detection speed of their method, contributing to the resilience of VPPs against cyber threats.

While these studies demonstrate FL's potential in CPS applications, they primarily focus on either detection accuracy or system efficiency in isolation, lacking comprehensive solutions for industrial condition monitoring. Our work addresses these limitations through key innovations: (1) an adaptive FL framework with dynamic model aggregation for sensor reliability, (2) integrated anomaly localization, and (3) a novel Weibull-based checkpointing mechanism. We validate these innovations on NASA bearings \cite{qiu2006wavelet} and Hydraulic systems \cite{helwig2015condition} datasets, demonstrating improved accuracy, efficiency, and robustness under node failures—advances not previously achieved in existing work.

%% file: 40-methods.tex
\section{Anomaly Detection Framework}
\label{sec:framework}

\begin{figure}[htbp]
    \centering
    \includegraphics[width=0.98\linewidth]{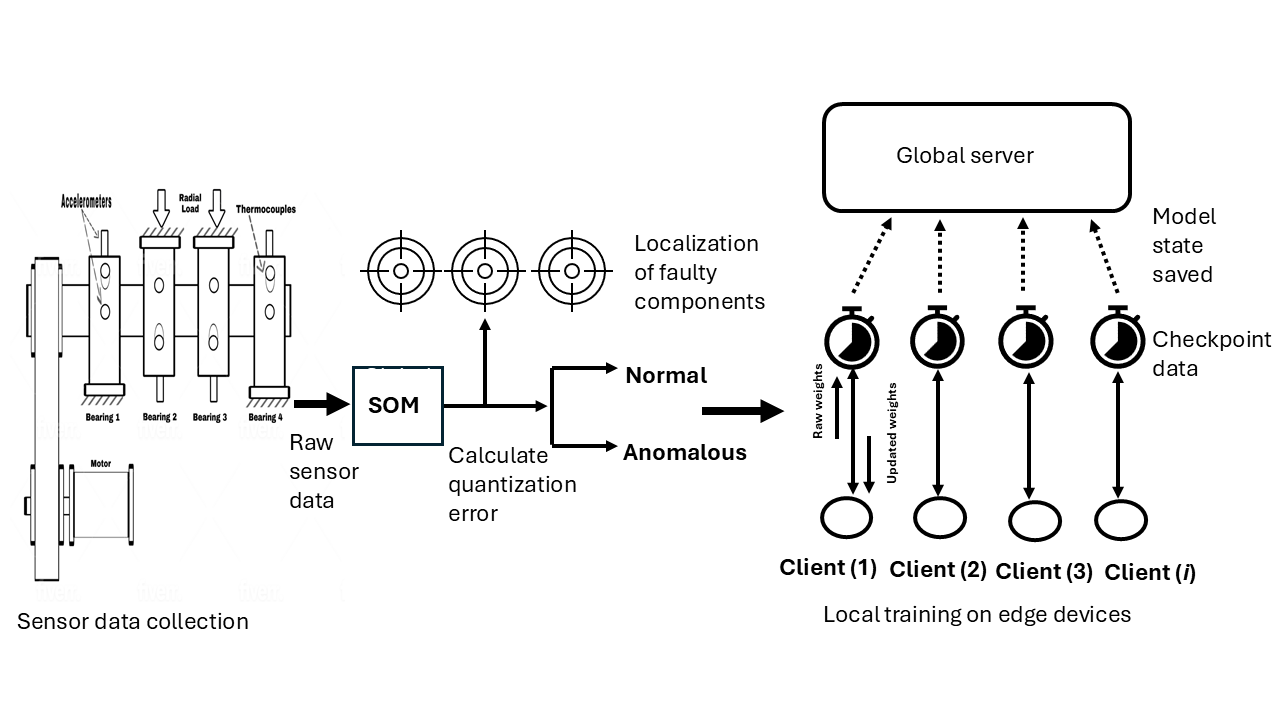}
    \caption{Framework architecture showing data flow from sensor collection through SOM-based detection to FL with adaptive checkpointing. The system processes raw sensor data to detect anomalies, localizes faulty components, and maintains model consistency across distributed clients.}
    \label{fig:workflow}
\end{figure}

This section introduces our framework for real-time condition monitoring in CPS environments. While our framework is designed for live industrial deployments, we validate its capabilities using established datasets that emulate real-world operational scenarios. Fig.~\ref{fig:workflow} illustrates how our framework orchestrates the flow of sensor data through multiple processing stages, from initial collection to FL analysis.

Our framework addresses three critical CPS challenges: computational burden in distributed environments, precise component-level anomaly localization, and system reliability under node failures. The workflow begins with sensor data collection and preprocessing, feeds into SOM-based anomaly detection, and culminates in FL across distributed clients. Each stage incorporates specific enhancements for industrial environments - from adaptive model aggregation based on sensor reliability to Weibull-based checkpointing for fault tolerance. The following subsections detail each component's implementation and their interactions within the overall system.

\subsection{Industrial Sensor Data Collection and Processing}
\label{sec:DataCollection}

Our framework's evaluation requires datasets that provide real industrial scenarios, document fault progression patterns, and offer opportunities to verify both detection and localization capabilities. We select two comprehensive industrial datasets that meet these criteria due to their ability to represent real-world operational conditions and capture detailed fault progression, making them well-suited for validating our anomaly detection and localization methods.

\begin{figure}[htbp]
    \centering     \vspace{-10pt}
    \includegraphics[width=0.25\textwidth]{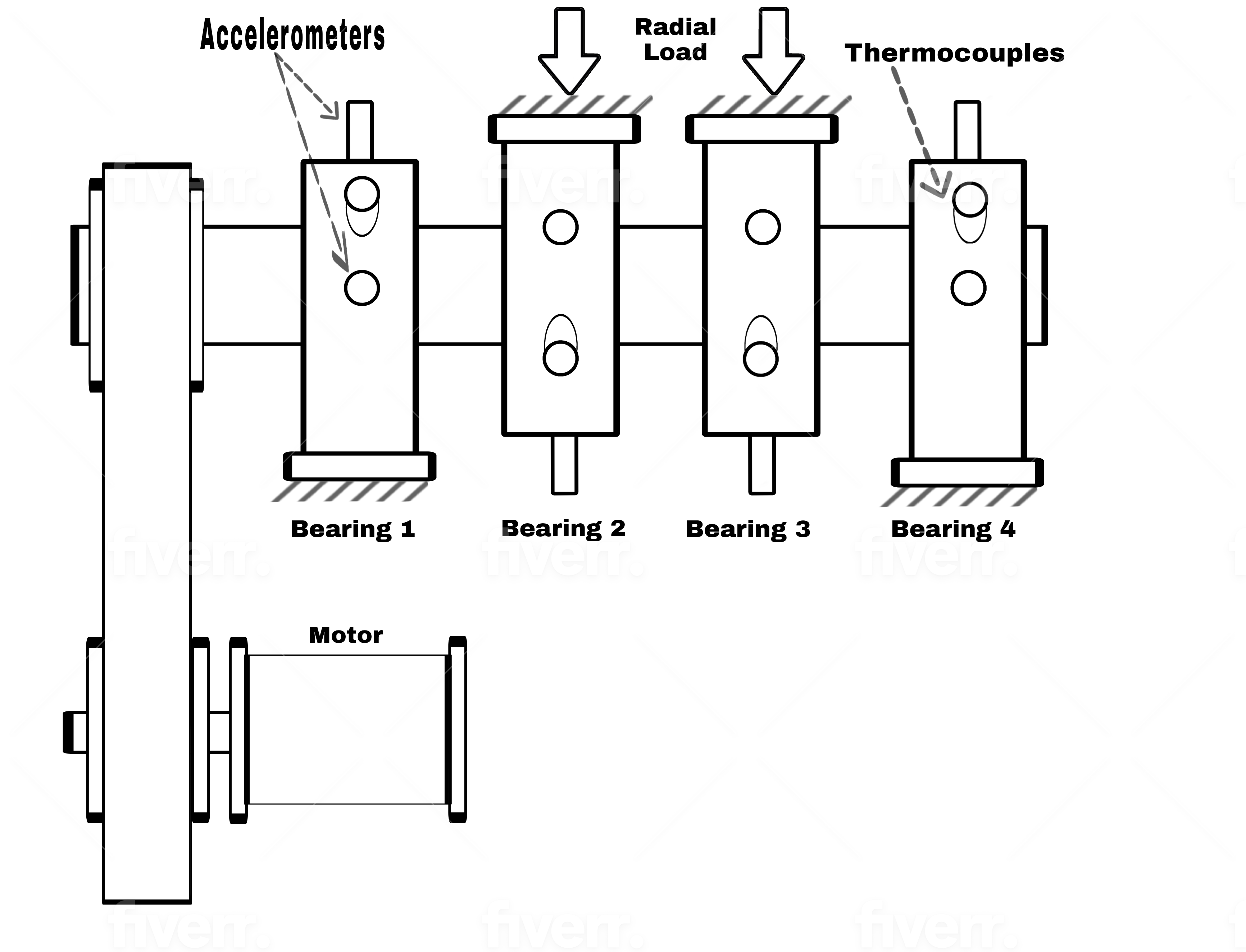} 
    \caption{Bearing test rig configuration showing sensor placement for vibration monitoring}\vspace{-10pt}
    \label{fig2}
\end{figure}

The NASA bearings dataset \cite{qiu2006wavelet} is generated using a specialized test rig with four Rexnord ZA-2115 double-row bearings mounted on a shaft operating at 2000 RPM under a 0.45 kg radial load. PCB 353B33 high-sensitivity quartz ICP accelerometers capture vibration data at 20 kHz, providing 20,480 observations per second. We utilize sets 1-3, each containing 984 files that progress from normal operation to failure, enabling validation of both detection accuracy and fault progression tracking. Fig.~\ref{fig2} shows the sensor configuration. The dataset's comprehensive bearing degradation records make it ideal for evaluating our approach's ability to identify early signs of failure and monitor the progression of faults.

The hydraulic systems dataset \cite{helwig2015condition} comprises 2205 instances from a test rig with primary working and cooling-filtration circuits. The system captures comprehensive measurements through multiple sensor types: pressure and motor power sensors operating at 100 Hz generating 6000 attributes each, volume flow sensors at 10 Hz producing 600 attributes each, and various monitoring sensors including temperature, vibration, and efficiency factors at 1 Hz yielding 60 attributes each. This dataset enables validation across various fault types, with labels indicating different operational states such as cooler conditions, valve states, and system stability. The variety of sensor data and fault labels make this dataset highly suitable for testing our framework's ability to localize anomalies across different components and operational conditions.

To ensure robust analysis, we implement a systematic preprocessing approach. The data undergoes normalization to a [0,1] range, stabilizing model performance across both datasets. We apply band-pass and low-pass filtering \cite{samoilov2024maximum} to remove noise and enhance signal integrity, followed by feature selection using permutation importance \cite{marfo2022condition} to optimize computational efficiency while maintaining detection accuracy. This preprocessed data forms the foundation for our anomaly detection mechanism, which employs SOM to identify deviations from normal operation.

\subsection{Anomaly Detection through Self-Organizing Maps}
\label{sec:SOM}

Our framework employs SOM \cite{kohonen1990self} for anomaly detection due to its ability to capture non-linear relationships in high-dimensional sensor data and provide quantifiable deviation measures. We initialize the SOM with a 50x50 neuron grid to balance computational complexity with detection accuracy, as this dimension effectively captures the feature space of our industrial datasets while maintaining reasonable training time.

The training process uses the first 60\% of each dataset as baseline data, representing normal operational conditions. To ensure numerical stability and consistent feature contribution, we normalize the input data with an epsilon adjustment: \(\mathbf{x}_{\text{normalized}} = \text{MinMaxScaler}(\mathbf{x}) + \epsilon\), where \(\epsilon = 1 \times 10^{-10}\). During training, each input vector \(\mathbf{x}_i\) is mapped to its best matching unit (BMU) by minimizing the distance between the input and neuron weight vectors. The quantization error, defined as \(e_i = \|\mathbf{x}_i - \mathbf{w}_{\text{BMU}}\|\), provides our primary metric for anomaly detection.

After 50 training iterations with a Gaussian neighborhood function, we establish an anomaly threshold based on the statistical distribution of training set quantization errors: \(\text{Threshold} = \mu_e + 3\sigma_e\), where \(\mu_e\) and \(\sigma_e\) represent the mean and standard deviation of baseline errors. This threshold derivation ensures robust anomaly detection by accounting for natural variations in the data while maintaining sensitivity to significant deviations. Any data point generating a quantization error above this threshold indicates a potential anomaly, enabling our framework to identify developing faults before they lead to system failure. While detecting anomalies provides crucial insights, localizing these issues to specific components enables targeted maintenance interventions.

\subsection{Component-Level Anomaly Localization}
\label{sec:AnomalyLocalization}

The precise identification of anomalous components enhances maintenance efficiency in CPS environments by enabling targeted interventions before failures occur. Our localization approach analyzes the spatial distribution of anomalies across sensors to pinpoint specific components requiring attention. For each sensor \(S_j\), we compute a cumulative anomaly score: \(C_j = \sum_{i=1}^{n} I(A_{ij} > \text{Threshold})\), where \(A_{ij}\) represents the anomaly score for sensor \(S_j\) at time \(t_i\), and \(I(\cdot)\) is an indicator function returning 1 when the score exceeds our detection threshold. This computation provides a quantitative measure of each component's contribution to system anomalies. The resulting distribution of \(C_j\) values enables our framework to rank components by their anomaly frequency, creating a prioritized maintenance schedule that optimizes resource allocation and minimizes system downtime. Having established our detection and localization approaches, we now detail how FL enables distributed implementation of these capabilities.

\subsection{FL Implementation}
\label{sec:FL}

Our FL architecture addresses critical limitations in traditional FL approaches when applied to CPS environments. Standard federated averaging faces three key challenges in industrial settings: it assumes uniform reliability across all clients, lacks mechanisms for handling sensor degradation, and treats all client updates with equal importance regardless of their detection performance. These limitations significantly impact anomaly detection accuracy in real-world industrial deployments where sensor reliability varies and node performance fluctuates over time.

To overcome these challenges, we implement a performance-aware client selection and aggregation strategy. Each client \(c_i\) maintains a local dataset \(X_i\) and model \(f_i\), training for \(e\) epochs to produce parameters \(w_{f_i} = f_i(X_i, e)\). Unlike traditional FL, our framework continuously evaluates client performance through three key metrics: anomaly detection accuracy on a validation set, sensor drift measurements, and historical prediction consistency. This evaluation enables our system to identify and prioritize high-performing clients, ensuring model updates originate from the most reliable data sources in the CPS network.

The global server employs our enhanced aggregation mechanism that builds upon \cite{mcmahan2017communication} by incorporating these performance metrics. The global model parameters \(w_g\) are computed as: \(w_g = \sum_{i=1}^{n} \alpha_i w_{f_i}\), where \(\alpha_i\) represents our adaptive weighting factor: \(\alpha_i = \frac{\beta_i \cdot \gamma_i \cdot \delta_i}{\sum_{j=1}^{n} \beta_j \cdot \gamma_j \cdot \delta_j}\). Here, \(\beta_i\) represents detection accuracy, \(\gamma_i\) captures sensor reliability, and \(\delta_i\) measures prediction stability for client \(i\). These parameters are estimated as follows: Detection accuracy ($\beta_i$) is detailed in Section~\ref{sec:Results}. Sensor reliability ($\gamma_i$) is quantified through $\gamma_i = \exp(-|\sigma_i - \sigma_{ref}|/\sigma_{ref})$ where $\sigma_i$ is the current sensor variance and $\sigma_{ref}$ is the reference variance from initial calibration. Prediction stability ($\delta_i$) is measured using $\delta_i = \frac{1}{1 + \text{var}(p_t - p_{t-1})}$ where $\text{var}(p_t - p_{t-1})$ is the variance of prediction changes over a sliding window. This multi-factor weighting approach ensures that clients demonstrating consistent performance and reliable sensor readings have a greater influence on the global model while automatically reducing the impact of degraded or unstable nodes. By dynamically adjusting these weights based on ongoing performance evaluation, our system maintains robust anomaly detection capabilities even as client conditions change over time. To maintain reliable operation of this distributed framework, we implement an adaptive fault tolerance mechanism detailed in the following subsection.

\subsection{Adaptive Fault Tolerance Mechanism}
\label{sec:FaultTolerance}

Traditional FL systems in CPS environments face significant challenges when handling node failures and client dropouts. Current approaches typically employ two inadequate recovery strategies: complete system restart or client reinitialization with the last known global weights. These methods result in substantial training disruptions and resource inefficiencies, particularly problematic in time-sensitive industrial monitoring applications.

Our framework addresses these limitations through an adaptive checkpointing mechanism based on failure probability modeling. Unlike standard checkpointing in deep learning frameworks, our approach dynamically adjusts save intervals based on system conditions. We model node failure probability using a Weibull distribution, chosen for its effectiveness in representing industrial system failure patterns: \(F(t) = 1 - \exp\left(-\left(\frac{t}{\lambda}\right)^k\right)\), where \(\lambda\) and \(k\) are scale and shape parameters derived from historical CPS failure data. The optimal checkpointing interval \(t_c^*\) balances overhead costs against recovery time through our cost function: \(C(t_c) = \frac{t_c}{T} + p_f(t_c) \cdot \frac{t_r}{T}\).

This formulation accounts for total computation time \(T\), recovery time \(t_r\), and failure probability \(p_f(t_c)\). By minimizing this cost function, we determine checkpoint intervals that adapt to changing system conditions, ensuring efficient recovery while minimizing overhead.

During operation, each client maintains state information including model parameters, optimization states, and training progress. Upon detecting a failure, our system leverages these checkpoints to restore the failed client's state and synchronize with the global model, eliminating the need for complete retraining. By minimizing training disruptions and adapting to system conditions, our adaptive fault tolerance mechanism ensures reliable model training across distributed nodes. This fault tolerance capability plays a crucial role in maintaining real-time anomaly detection and system health monitoring, thereby enhancing the robustness of our FL framework in CPS environments.

%% file: 50-results.tex
\section{Results}
\label{sec:Results}

This section evaluates our framework's effectiveness in addressing the key challenges of anomaly detection in CPS environments. We assess our approach's ability to maintain detection accuracy under varying sensor reliability, handle node failures, and provide efficient anomaly localization. Our experiments simulate a distributed CPS environment to validate the framework under realistic operational conditions.

\subsection{Experimental Setup}
\label{sec:Setup}

We implement our framework using PyTorch for the FL components, with experiments conducted on an Intel i9-12900HK processor with 32GB RAM. Our evaluation environment simulates 10 distributed clients processing sensor data from the NASA Bearings and Hydraulic System Datasets, reflecting real-world CPS deployments where edge devices operate across different locations.

The neural network architecture in each client comprises four fully connected layers (128, 64, 32 neurons) with ReLU activation functions and a 40\% dropout rate. The dropout rate is specifically chosen to prevent overfitting by randomly deactivating 40\% of neurons during training, which is particularly important given the high-dimensional nature of sensor data. We use the Adam optimizer with a learning rate of 0.001, which offers a balance between convergence speed and stability. The batch size is set to 512 to efficiently process large amounts of sensor data, allowing the model to learn effectively from diverse data points while maintaining computational efficiency. Early stopping is incorporated to prevent overfitting, ensuring that training halts once the model performance plateaus on the validation set. 

For a comprehensive evaluation, we include state-of-the-art methods such as FedAvg \cite{mcmahan2017communication}, ACFL \cite{yan2023criticalfl}, and FedL2P \cite{lee2024fedl2p} as our baselines.

\subsubsection{Evaluation Metrics}
\label{sec:PerformanceMetrics}

\textit{(i) Detection accuracy}: Measures the proportion of correct predictions, offering a straightforward view of model performance. However, it may be sensitive to class imbalance, particularly in datasets where anomalies are rare.

\textit{(ii) AUC-ROC}: Evaluates model performance across different classification thresholds, crucial for imbalanced datasets like ours. It quantifies the model's ability to distinguish between normal and anomalous instances and is calculated as: $\text{AUC-ROC} = \int_{0}^{1} \text{TPR}(\text{FPR}^{-1}(x)) \, dx$.

\subsection{Framework Validation}
\label{sec:validation}

\begin{figure}[h]
    \centering
    \includegraphics[width=0.95\linewidth]{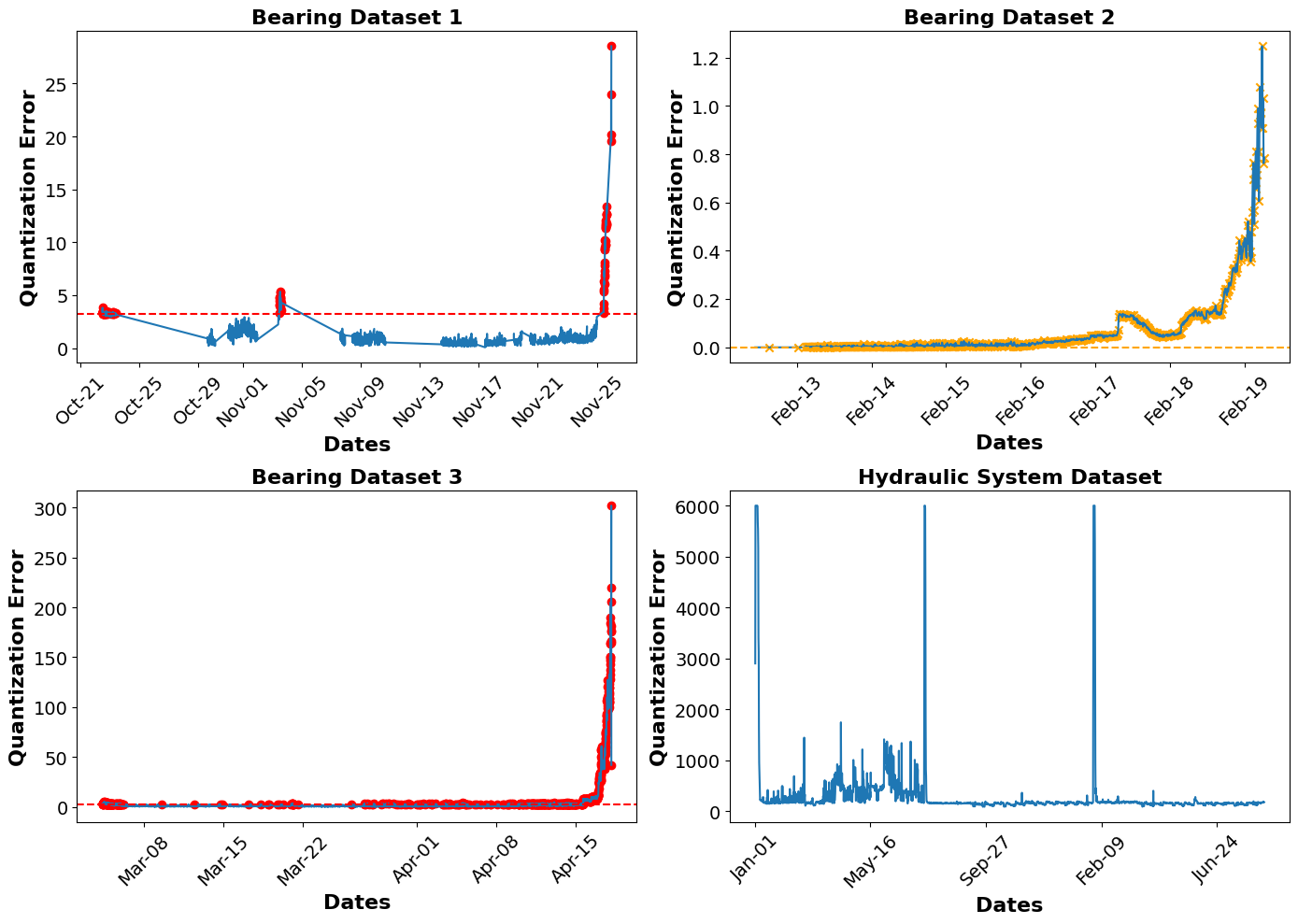}
    \caption{SOM-based anomaly detection across datasets showing quantization error progression: Bearing Dataset 1 (top-left) exhibits gradual degradation, Bearing Dataset 2 (top-right) shows early-stage deterioration, Bearing Dataset 3 (bottom-left) demonstrates abrupt failure, and Hydraulic System (bottom-right) reveals multi-component anomaly patterns.}
    \label{fig:anomaly}
\end{figure}

Our framework demonstrates effective anomaly detection and localization capabilities in CPS environments through its enhanced FL components. Fig.~\ref{fig:anomaly} shows the SOM-based detection results across different operational scenarios, revealing our framework's ability to identify various failure patterns.

The detection results reveal distinct failure patterns across datasets. Bearing Dataset 1 shows gradual degradation beginning November 3, with periodic spikes culminating in severe deterioration by November 25. Bearing Dataset 2 demonstrates early-stage fault detection around February 17, where the quantization error transitions from stable operation to elevated values with total degradation after February 19. Bearing Dataset 3 captures an abrupt failure scenario characterized by a sharp increase in quantization error after April 15. The Hydraulic Systems Dataset exhibits distinct temporal patterns with significant anomaly spikes, validating our framework's ability to handle complex multi-component systems.

\begin{figure}[h]
    \centering
    \includegraphics[width=0.97\linewidth]{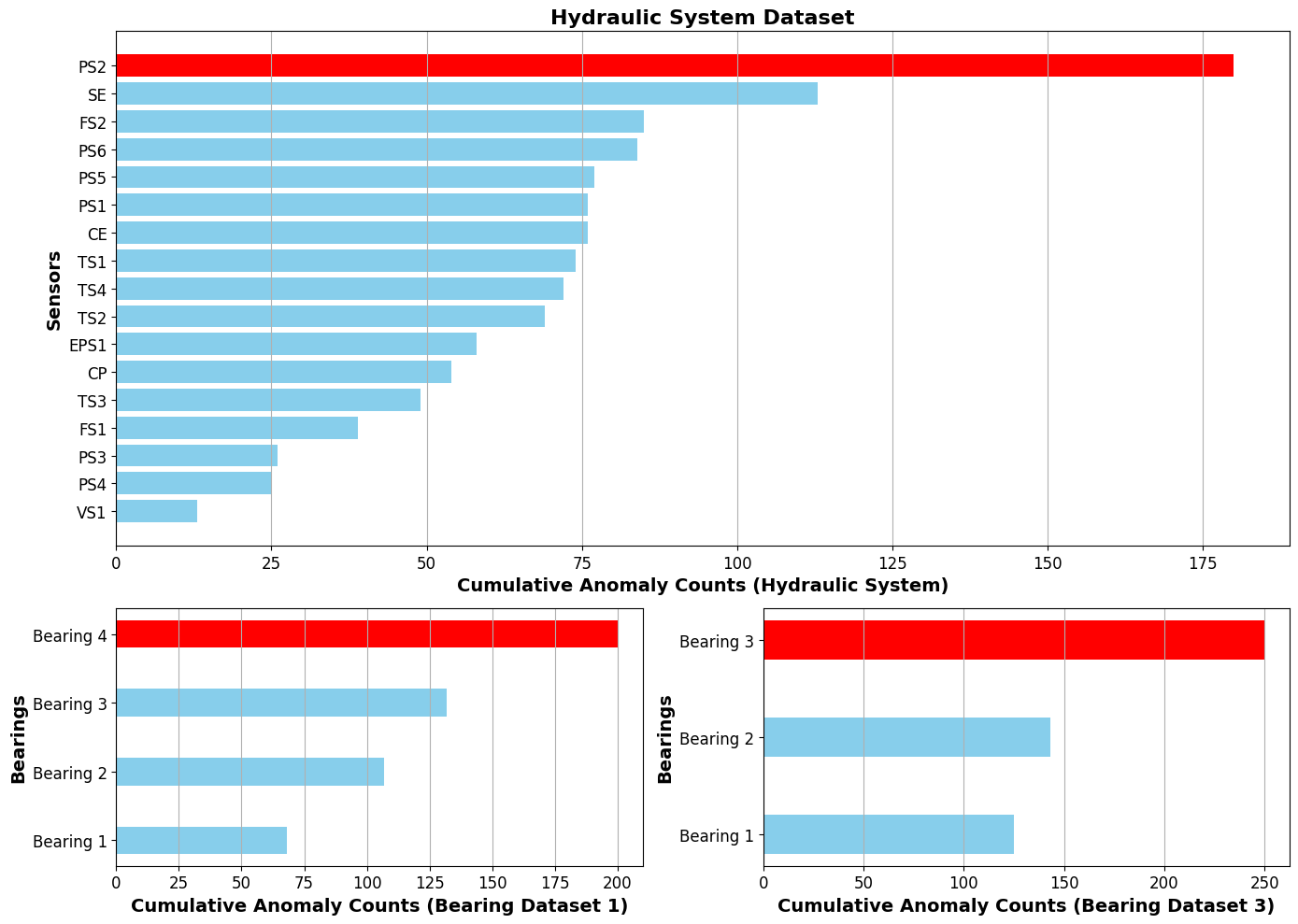}
    \caption{Component-level anomaly localization showing cumulative anomaly counts: Bearing Datasets 1 and 3 (bottom) identify critical components, while the Hydraulic System (top) reveals sensor-specific degradation patterns. Localization results for Bearing Dataset 2 align with previous findings \cite{marfo2022condition}.}
    \label{fig:localization}
\end{figure}

Our framework's anomaly localization capability enables precise identification of problematic components. Fig.~\ref{fig:localization} demonstrates this through cumulative anomaly counts across datasets. In the bearings case, our approach consistently identifies critical components, with Bearing 3 showing significantly higher anomaly counts in both datasets. The Hydraulic System analysis reveals Pressure Sensor 2 (PS2) as the primary concern, with elevated counts compared to other sensors, enabling targeted maintenance planning.

Quantitative comparison with existing FL approaches in Table~\ref{tab:method_comparison} demonstrates our framework's superior performance. The improved accuracy and AUC-ROC scores result from our adaptive model aggregation strategy, which dynamically weights client contributions based on sensor reliability. While FedAvg suffers from uniform client treatment and ACFL and FedL2P show limited improvement, our framework achieves consistently high performance through its CPS-specific optimizations.

\begin{table}[h]
\centering
\caption{Performance comparison across FL methods with 10 clients and 15 rounds demonstrating effectiveness of our adaptive approach.}
\label{tab:method_comparison}
\begin{tabular}{l|c|c|c}
\hline
\textbf{Method} & \textbf{Accuracy (\%)} & \textbf{AUC-ROC} & \textbf{Time (s)} \\ \hline
\multicolumn{4}{c}{\textbf{Bearings Dataset}} \\ \hline
Proposed        & \textbf{99.5}          & \textbf{0.995}    & \textbf{35}     \\
ACFL            & 95.8                   & 0.962             & 52              \\
FedL2P          & 93.2                   & 0.945             & 48              \\
FedAvg          & 90.1                   & 0.912             & 65              \\ \hline
\multicolumn{4}{c}{\textbf{Hydraulic Systems Dataset}} \\ \hline
Proposed        & \textbf{98.3}          & \textbf{0.981}    & \textbf{42}     \\
ACFL            & 94.7                   & 0.953             & 58              \\
FedL2P          & 92.5                   & 0.934             & 55              \\
FedAvg          & 89.4                   & 0.901             & 71              \\
\hline
\end{tabular}\vspace{-1em}
\end{table}

These results directly address RQ1 by demonstrating our framework's ability to achieve reliable condition monitoring in CPS environments through enhanced FL components. The combination of accurate anomaly detection, precise component localization, and adaptive model aggregation enables robust real-time monitoring across diverse industrial settings.

\subsection{Scalability and Fault Tolerance Evaluation}
\label{sec:scalability_fault}

\begin{figure}[h]
    \centering
    \includegraphics[width=0.97\linewidth]{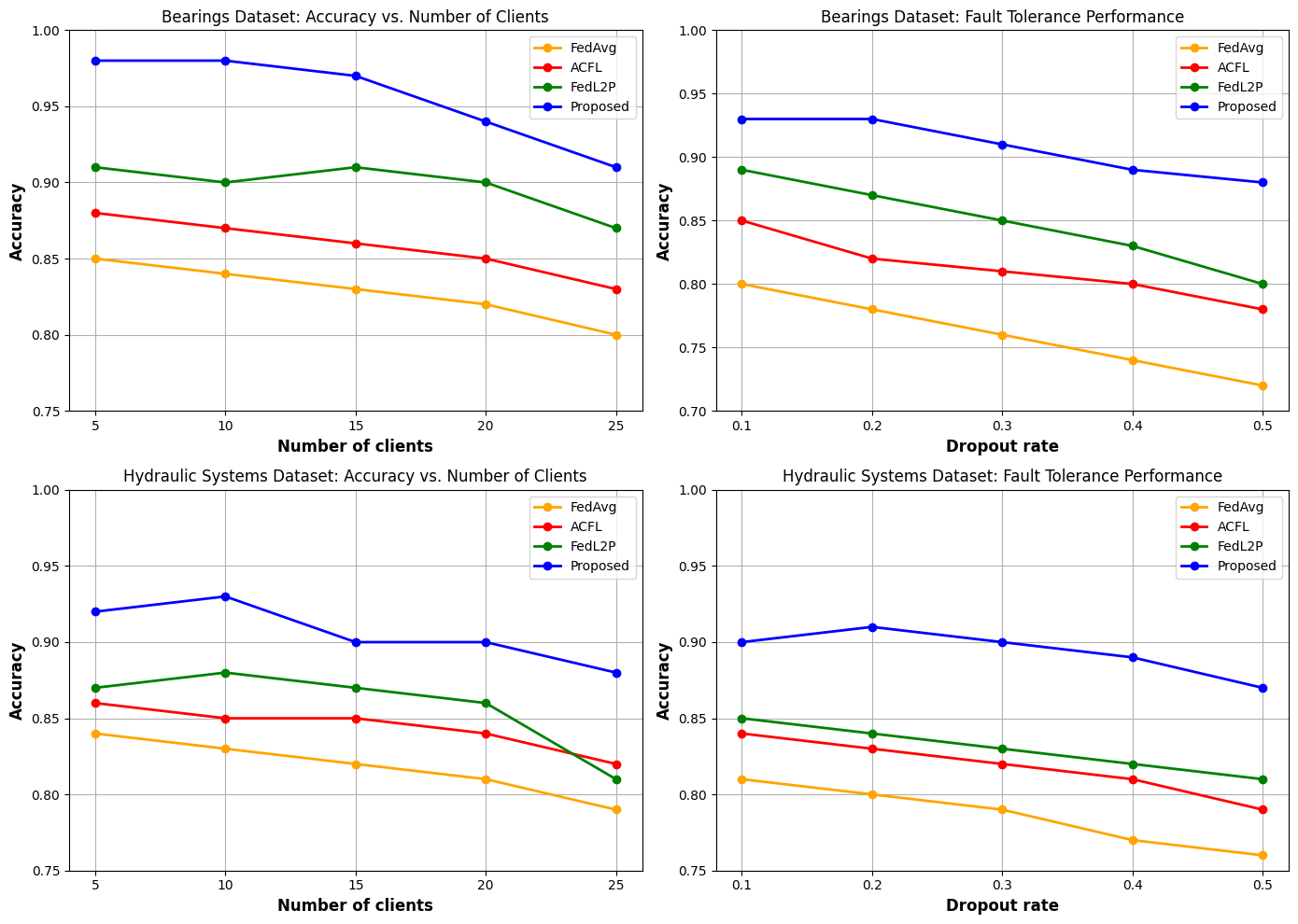}
    \caption{Performance analysis showing accuracy versus number of clients (left) and dropout rates (right) for both datasets. Our framework maintains superior detection capability through adaptive aggregation and Weibull-based checkpointing.}
    \label{fig:scalability}
\end{figure}

Our framework's effectiveness in real-world CPS deployments depends on both its ability to scale across distributed nodes and maintain performance under node failures in 15 rounds. Fig.~\ref{fig:scalability} demonstrates these capabilities across both datasets.

For the Bearings Dataset, our framework maintains detection accuracy above 90\% as client numbers increase to 25, while other approaches show significant degradation beyond 15 clients. This superior scaling stems from our adaptive model aggregation strategy, which intelligently weights client contributions based on sensor reliability. The framework's resilience is particularly evident under increasing dropout rates, where it maintains above 85\% accuracy even at 0.5 dropout rate. This robustness results from our Weibull-based checkpointing mechanism, which adaptively saves training states based on predicted failure patterns.

The Hydraulic Systems Dataset results further validate these capabilities in a more complex environment. Our framework maintains accuracy above 85\% across varying client configurations and demonstrates notably slower performance degradation under node failures compared to baseline methods. While FedAvg's performance drops sharply due to its uniform client treatment and lack of fault tolerance, and ACFL and FedL2P show moderate decline due to their limited adaptation capabilities, our framework's combination of adaptive aggregation and strategic checkpointing ensures consistent performance.

These results directly address RQ2 and RQ3, demonstrating our framework's ability to maintain reliable operation at scale while handling node failures effectively. The superior performance derives from our framework's key innovations: adaptive aggregation for varying sensor reliability and Weibull-based checkpointing that optimizes fault tolerance based on operational patterns. This combination enables robust operation in practical CPS deployments where both scaling requirements and node failures are common challenges.

\subsection{Statistical Validation}  
\label{sec:stats}

To rigorously validate our framework's performance advantages, we conduct a statistical analysis comparing our approach against state-of-the-art FL methods (FedAvg, ACFL, and FedL2P). We employ the Mann-Whitney U test to evaluate the significance of performance differences in AUC-ROC scores across all experimental configurations. The Mann-Whitney U test, a non-parametric test, is chosen due to its robustness in comparing distributions without assuming normality, which makes it particularly suitable for our detection accuracy measurements.

The null hypothesis (\(H_0\)) in our analysis is that there is no difference between the AUC-ROC values produced by our proposed method and the baseline methods (FedAvg, ACFL, and FedL2P). The alternative hypothesis (\(H_1\)) is that the AUC-ROC values produced by our method are significantly greater than those produced by the baselines, indicating improved performance. We conduct the tests at a significance level of \(\alpha = 0.05\). Table~\ref{tab:stat_significance} presents the results of the Mann-Whitney U test, demonstrating the statistical significance of our framework's performance improvements across both datasets.

The extremely low \(p\)-values (\(p < 0.05\)) across all comparisons allow us to reject the null hypothesis, confirming that our framework's performance improvements are statistically significant. This significance stems from our framework's key innovations: the adaptive model aggregation strategy better handles varying sensor reliability, while the dynamic node selection mechanism optimizes client participation. These advantages manifest consistently across both datasets, with particularly strong statistical significance in comparisons against state-of-the-art FL methods (FedAvg, ACFL, and FedL2P), which lack these CPS-specific optimizations. The results reinforce our framework's effectiveness in addressing the unique challenges of industrial CPS environments through its enhanced FL components.

\begin{table}[h]
\centering
\caption{Mann-Whitney U test results demonstrating the statistical significance of our framework's performance improvements across both datasets.}
\label{tab:stat_significance}
\begin{tabular}{l|c|c|c}
\hline
\textbf{Comparison} & \textbf{Dataset} & \textbf{U Statistic} & \textbf{\textit{p}-value} \\
\hline
Proposed vs. FedAvg & Bearings (1-3) & 11678.0 & 5.12e-17 \\
Proposed vs. ACFL & Bearings (1-3) & 12045.0 & 2.85e-17 \\
Proposed vs. FedL2P & Bearings (1-3) & 11924.0 & 3.78e-16 \\
\hline
Proposed vs. FedAvg & Hydraulic & 11523.0 & 4.89e-17 \\
Proposed vs. ACFL & Hydraulic & 11892.0 & 3.12e-17 \\
Proposed vs. FedL2P & Hydraulic & 11756.0 & 4.23e-16 \\
\hline
\end{tabular}\vspace{-1em}
\end{table}

%% file: 55-discussion.tex
\section{Discussion}
\label{sec:Discussion}

Our results demonstrate the effectiveness of the enhanced FL framework for reliable condition monitoring in CPS environments. Key innovations—adaptive model aggregation, dynamic node selection, and Weibull-based checkpointing—drive its superior performance. The framework achieves 99.5\% AUC-ROC on the Bearings Dataset and 98.3\% on the Hydraulic Systems Dataset, outperforming ACFL by 3.7\% and 3.6\%, respectively. Adaptive model aggregation intelligently weights client contributions, significantly improving detection accuracy compared to FedAvg (90.1\%). Dynamic node selection optimizes resource use, resulting in 2x faster execution, while Weibull-based checkpointing enhances fault tolerance. These features are crucial for maintaining high accuracy and minimizing disruptions in industrial settings. The framework's practical value lies in reducing downtime and optimizing resource allocation, leading to improved efficiency and cost savings for industries.

However, this work assumes historical failure data for Weibull modeling, which may not always be available. Additionally, further validation across more CPS environments is needed to enhance generalizability. Future work should focus on developing an online adaptation of Weibull parameters and anomaly detection framework, supporting diverse sensor types and failure modes, and reducing communication overhead for better scalability in larger deployments. Overall, the proposed framework demonstrates significant potential for FL in CPS, offering enhanced accuracy, scalability, and resilience with broad industrial applicability.

%% file: 60-conclusion.tex
\section{Conclusion}
\label{sec:Conclusion}
This paper presents an enhanced FL framework for CPS condition monitoring, integrating adaptive model aggregation, dynamic node selection, and Weibull-based checkpointing. Evaluation on NASA Bearing and Hydraulic System Datasets demonstrates superior performance with 99.5\% AUC-ROC accuracy, outperforming existing approaches like FedAvg, ACFL, and FedL2P. Future work will focus on reducing computational overhead and extending the framework to broader industrial CPS applications where continuous monitoring is critical.\vspace{-1em}